\documentclass[letterpaper, 10pt, conference]{ieeeconf}

\usepackage[utf8]{inputenc}
\usepackage[T1]{fontenc}
\usepackage{acro}
\usepackage{amsfonts}
\usepackage{amsmath}
\usepackage[keeplastbox, nocheckfootnote]{flushend}
\usepackage{multirow}
\usepackage{pgfplots}
\usepackage{pgfplotstable}
\usepackage[separate-uncertainty=true]{siunitx}
\usepackage{subfig}
\usepackage{tikz}
\usepackage{url}
\usepackage{xcolor}

\usetikzlibrary{calc, shapes.misc, arrows, positioning, 3d, scopes, patterns}

\pgfplotsset{compat=1.15}
\newcommand*\circled[1]{\tikz[baseline=(char.base)]{\node[shape=circle, draw, inner sep=2pt] (char) {#1};}}

\DeclareAcronym{BCE}{short=BCE, long=binary crossentropy}
\DeclareAcronym{DoF}{short=DoF, long=degree of freedom, long-plural-form=degrees of freedom}
\DeclareAcronym{FCNN}{short=FCNN, long=fully-convolutional neural network}
\DeclareAcronym{NN}{short=NN, long=neural network}
\DeclareAcronym{MDP}{short=MDP, long=Markov decision process}
\DeclareAcronym{RL}{short=RL, long=reinforcement learning}
\DeclareAcronym{PPH}{short=PPH, long=picks per hour}
\DeclareAcronym{TCP}{short=TCP, long=tool center point}

\DeclareMathOperator*{\argmax}{arg\,max}

\title{\LARGE \bf
    Robot Learning of 6\,DoF Grasping \\
    using Model-based Adaptive Primitives
}

\author{
    Lars Berscheid$^{1}$, Christian Friedrich$^{2}$, and Torsten Kröger$^{1}$%
    \thanks{
        $^{1}$Karlsruhe Institute of Technology (KIT), Germany
	    {\tt\small \{lars.berscheid, torsten\}@kit.edu}}
	\thanks{
	    $^{2}$SCHUNK GmbH \& Co. KG, Lauffen am Neckar, Germany
	    {\tt\small christian.friedrich@de.schunk.com}
	}
}

\IEEEoverridecommandlockouts

\begin{document}

\definecolor{myblue}{RGB}{38, 70, 83}
\definecolor{myred}{RGB}{231, 111, 81}
\definecolor{mygreen}{RGB}{42, 157, 143}
\definecolor{myyellow}{RGB}{233, 196, 106}

\maketitle

\begin{abstract}
Robot learning is often simplified to planar manipulation due to its data consumption. Then, a common approach is to use a \ac{FCNN} to estimate the reward of grasp primitives. In this work, we extend this approach by parametrizing the two remaining, lateral \acp{DoF} of the primitives. We apply this principle to the task of 6\,\ac{DoF} bin picking: We introduce a model-based controller to calculate angles that avoid collisions, maximize the grasp quality while keeping the uncertainty small. As the controller is integrated into the training, our hybrid approach is able to learn about and exploit the model-based controller. After real-world training of \num{27000} grasp attempts, the robot is able to grasp known objects with a success rate of over \SI{92}{\%} in dense clutter. Grasp inference takes less than \SI{50}{ms}. In further real-world experiments, we evaluate grasp rates in a range of scenarios including its ability to generalize to unknown objects. We show that the system is able to avoid collisions, enabling grasps that would not be possible without primitive adaption.
\end{abstract}

\section{INTRODUCTION}

Grasping is a fundamental task throughout robotics, as it enables further manipulation and interaction with the robots environment. In many of today's and future applications like pick-and-place tasks in manufacturing, large-scale warehouse automation or service robotics, reliable grasping of unknown objects remains challenging. In particular, \textit{bin picking} is the task of grasping objects out of unsystematic environments like a randomly filled bin. Several difficulties in comparison to general robotic grasping are highlighted: Noisy sensor data, an obstacle-rich environment, and a usually contact-rich task execution. State-of-the-art industrial solutions require an object model, which is localized within a point cloud via pose estimation. Usually, it is then grasped at pre-defined points \cite{siciliano_springer_2016}. However, model-free solutions for unknown objects continue to pose a challenge.

Robot learning has great potential for manipulation tasks. As data-driven methods have the intrinsic capability to generalize, a universal grasping controller for many objects and scenarios might be within reach. As the data consumption remains the practical bottleneck for learning methods, the action space is frequently simplified to reduce costly exploration. Oftentimes, planar manipulation with 4\,\acp{DoF} is used \cite{kalashnikov_2018_qt, zeng_learning_2018, berscheid_shifting_2019, mahler_dex-net_2017}. In general, this is a powerful yet efficient simplification regarding bin picking, in particular when combined with a \acf{FCNN} for calculating grasp quality scores. Still, planar grasping leads to a restriction of possible poses, in particular regarding collisions between the gripper and the bin.

To mitigate this problem, we keep the learned part of planar grasping and add a model-based controller for calculating the two lateral angles as the remaining \acp{DoF}. As our key contribution, we introduce this novel hybrid approach for learning 6\,\ac{DoF} grasping. We adapt the orientation of the grasping primitives using a simple model-based controller and integrate this adaption within the real-world training cycle. We evaluate our proposed approach on 6\,\ac{DoF} grasping in real-world experiments with over \SI{120}{h} of training time.

\begin{figure}[t]
	\center
\begin{tikzpicture}
    \node[anchor=south west, inner sep=0] (overall) at (0,0) {\includegraphics[trim=30 0 40 0, clip, width=0.78\linewidth]{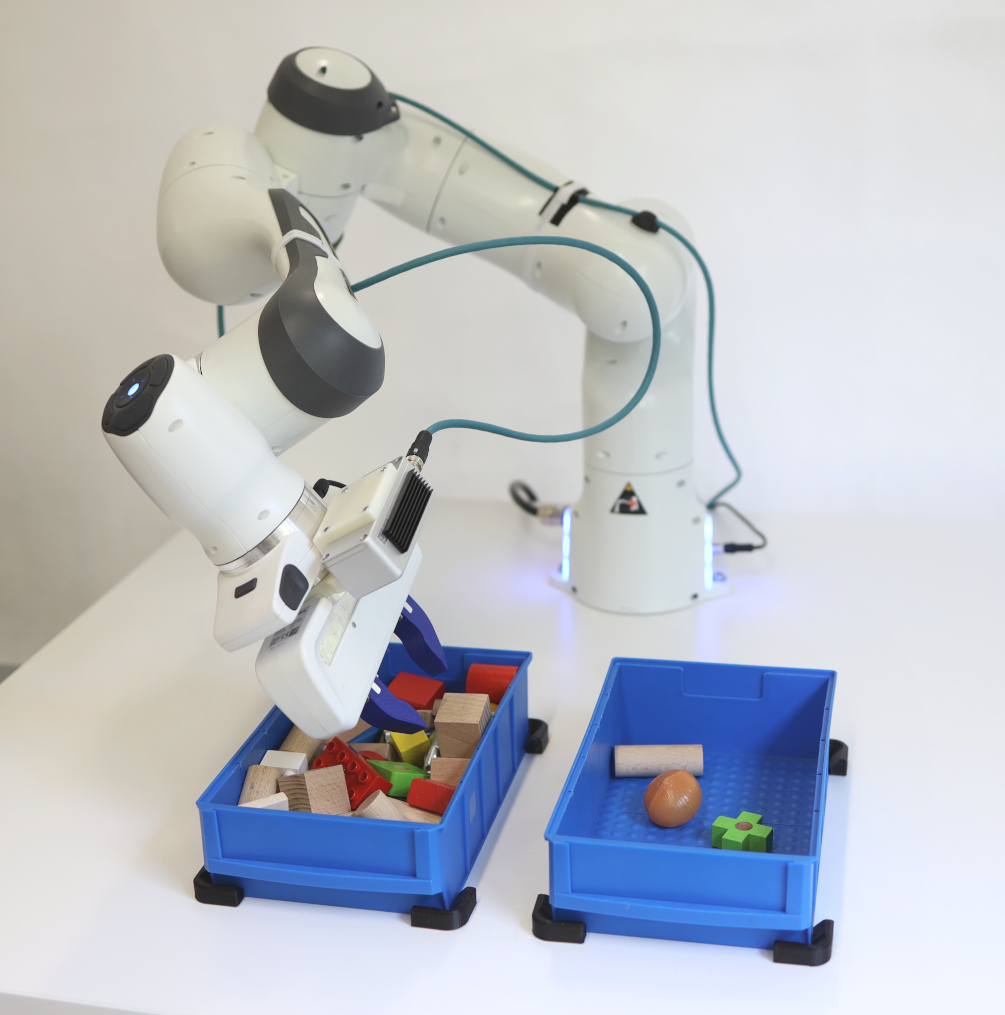}};
	\begin{scope}[x={(overall.south east)}, y={(overall.north west)}]
		\node[] at (0.89, 0.65) {\circled{1}};
		\draw [->] (0.863, 0.628) to (0.66, 0.44);

		\node[] at (0.11, 0.375) {\circled{2}};
		\draw [->] (0.144, 0.383) to (0.36, 0.44);
		
		\node[] at (0.11, 0.24) {\circled{3}};
		\draw [->] (0.142, 0.254) to (0.34, 0.34);
		
		\node[] at (0.89, 0.42) {\circled{4}};
		\draw [->] (0.876, 0.389) to (0.78, 0.19);
	\end{scope}
\end{tikzpicture}

	\caption{The robot avoids a possible collision with the bin by adapting the two lateral degrees of freedom by a model-based controller. The remaining planar grasping is learned in a self-supervised manner from real-world experiments. We evaluate our hybrid approach using the task of bin picking. Our system includes a robotic arm (1), a RGBD stereo camera (2), a two-finger gripper (3), and two bins with a multitude of objects (4).}
	\label{fig:overall-system}
\end{figure}
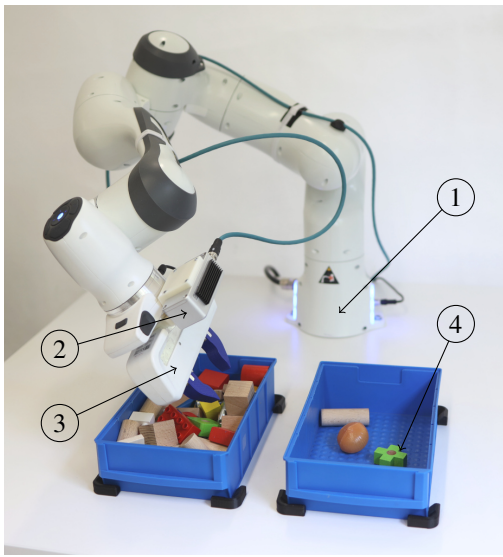

\section{RELATED WORK}

Grasping is a well-researched area within robotics. Bohg et al.~\cite{bohg_data-driven_2014} differentiate grasping of known, familiar and unknown objects. Early research has seen great progress for grasping known objects with object-model-based approaches. Here, grasp synthesis was based on analytical constructions of force-closure grasps \cite{ferrari_planning_1992}. Data-driven approaches usually \textit{sample} and \textit{rank} possible grasps \cite{miller_graspit!_2004}. Without object models or high-quality scans however, grasping remains challenging with classical methods. \\

\textbf{Robot learning} has shown great progress in recent years. Due to its ability to generalize, many approaches can easily be applied to unknown objects. For the task of grasping, we find that two major approaches have emerged here: First, an \textit{end-to-end} approach using a velocity-like control of the end-effector \cite{levine_learning_2016, kalashnikov_2018_qt, gualtieri2018learning}. Although this is a very powerful and expressive approach, it requires a lot of data due its time-dependent, high-dimensional action space. Second, a single-step approach using \textit{manipulation primitives} \cite{berscheid_improving_2019}. Here, a controller decides \textit{where} to apply \textit{which} pre-defined primitive. Combined with a \acf{FCNN} as an action-value estimator, this approach has been applied to a wide range of tasks including grasping \cite{berscheid_improving_2019, mahler_dex-net_2017}, pre-grasping manipulation \cite{zeng_learning_2018, berscheid_shifting_2019} or pick-and-place \cite{berscheid2020self}. As the primary data source, learned approaches uses either (a) real world measurements \cite{berscheid_improving_2019, pinto_supersizing_2016}, (b) analytical metrics \cite{mahler_dex-net_2017}, (c) simulation \cite{bousmalis_using_2017, quillen_deep_2018}, or (d) human annotations \cite{zeng2018robotic}. As exploration and inference scale exponentially with the action space, all but the latter simplify grasping to a \textit{planar} problem. \\

For data-driven or hybrid \textbf{6\,\ac{DoF} grasping}, simulation and analytical metrics have been used so far. In this context, ten Pas et al.~\cite{ten2017grasp} used a (mostly) model-based grasp synthesis. For the grasp evaluation, a convolutional \ac{NN} maps multiple depth images from different viewpoints around the grasp pose to a quality score. To avoid the costly rendering of the depth images, Qin et al.~\cite{qin2020s4g} proposed to learn analytical grasp-metrics directly from point-clouds. The grasp evaluation makes use of the PointNet++ architecture to learn a grasp measure from 3D spatial relations. Mousavian et al.~\cite{mousavian20196} introduced a grasp sample \ac{NN} based on a variational autoencoder and simulated training data. Subsequently, an iterative grasp refinement step selects and improves grasps taking the point-cloud as input. In contrast, Schmidt et al.~\cite{schmidt2018grasping} chooses a single depth image as the state space. Here, a convolutional \ac{NN} is trained to predict the 6~\ac{DoF} grasp pose as a regression problem. Due to its non-stochastic and single-modal nature, this approach works only for isolated objects. In this regard, our work focuses on dense clutter for bin picking scenarios. We also make use of model-based grasp sampling, however learn the final grasp evaluation in \textit{real-world} training.

\section{A HYBRID APPROACH FOR GRASPING}

We introduce our algorithm in the notation of \ac{RL}, however limited to a single action step. Then, the underlying \ac{MDP} is defined by the tuple $(\mathcal{S}, \mathcal{A}, r)$ with the state space $\mathcal{S}$, the action space $\mathcal{A}$, and the reward function $r$. Let $\mathcal{M}$ be a set of pre-defined manipulation primitives. Each action $a$ is then defined by applying a manipulation primitive $m$ at a pose $p \in \mathit{SE}(3)$.

\subsection{Action and State Spaces}

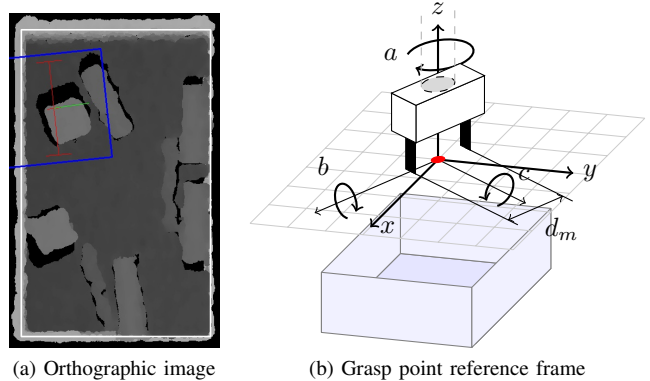
\begin{figure}[t]
	\centering
	\vspace{0.8mm}

\subfloat[Orthographic image]{
	\centering
	\includegraphics[trim=85 45 50 48, clip, width=0.25\textwidth, angle=90]{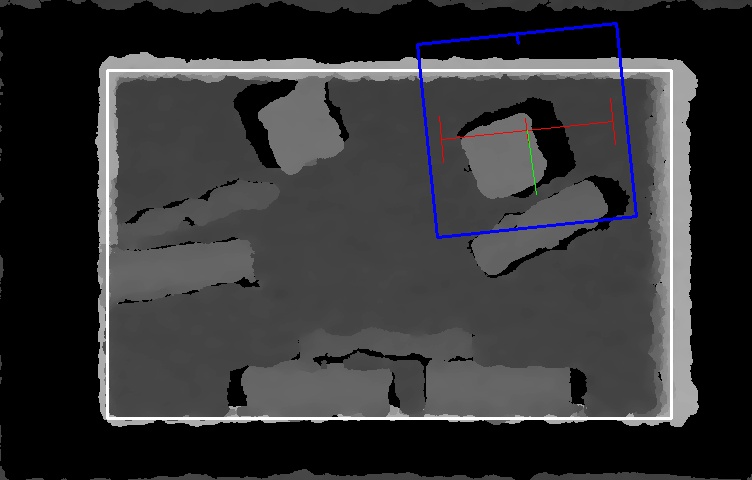}
	\label{fig:example-depth-image}
}
	~
\subfloat[Grasp point reference frame]{
	\centering
\begin{tikzpicture}[x={(-0.5cm, -0.5cm)}, y={(1cm, -0.1cm)}, z={(0cm, 1cm)}, scale=1]
	\definecolor{bincolor}{RGB}{80, 80, 255}
	\tikzstyle{facestyle} = [fill=bincolor!8, draw=gray, thin, line join=round]
	\tikzstyle{ann} = [black, inner sep=1pt]
	\tikzstyle{coordaxis} = [black, thick, ->]
	
	\begin{scope}[canvas is yx plane at z=0.2]
		\path[facestyle, fill=bincolor!15] (-0.5,0) rectangle (1.5,2);
	\end{scope}
	
	\begin{scope}[canvas is zy plane at x=0]
		\path[facestyle] (0.2,-0.5) rectangle (1,1.5);
	\end{scope}
	
	\begin{scope}[canvas is zx plane at y=-0.5]
		\draw[facestyle] (0.2,0) rectangle (1,2.1);
	\end{scope}
	
	\begin{scope}[canvas is zy plane at x=2.1]
		\path[facestyle] (0.2,-0.5) rectangle (1,1.5);
	\end{scope}
	
	\begin{scope}[canvas is zx plane at y=1.5]
		\draw[facestyle] (0.2,0) rectangle (1,2.1);
	\end{scope}
	
	\begin{scope}[canvas is xy plane at z=1.5]
		\draw[step=5mm, very thin, gray!40] (-1.5,-1.5) grid (2,2);
	\end{scope}
	
	\draw[coordaxis] (0,0,1.5) -- +(1.8,0,0) node[right] {$x$};
	\draw[coordaxis] (0,0,1.5) -- +(0,1.8,0) node[right] {$y$};
    \draw[coordaxis] (0,0,1.5) -- +(0,0,1.8) node[above] {$z$};
    
    \draw[coordaxis, <-] (0.2,-0.2,2.8) arc(-20:280:.4) node[left] {$a$};

	\begin{scope}[shift={(0, 0, 2.0)}, rotate around z=-28]
		\draw[black, fill=black] (0.4,-0.1,0) -- ++(0,0,-0.5) -- ++(0,0.2,0) -- ++(0,0,0.5) -- cycle;
		\draw[black, fill=black] (-0.4,-0.1,0) -- ++(0,0,-0.5) -- ++(0,0.2,0) -- ++(0,0,0.5) -- cycle;
		
		\draw[black, fill=white] (-0.5,0.25,0) -- ++(1,0,0) -- ++(0,0,0.5) -- ++(-1,0,0) -- cycle;
	    \draw[black, fill=white] (0.5,-0.25,0) -- ++(0,0.5,0) -- ++(0,0,0.5) -- ++(0,-0.5,0) -- cycle;
	    \draw[black, fill=white] (-0.5,-0.25,0.5) -- ++(0,0.5,0) -- ++(1,0,0) -- ++(0,-0.5,0) -- cycle;
	    \draw[black, thin, fill=gray!30, dashed] (0,0,0.5) circle (0.2);
	    \draw[gray!60, thin, dashed] (0.173,-0.1,0.5) -- ++(0,0,1);
	    \draw[gray!60, thin, dashed] (-0.173,0.1,0.5) -- ++(0,0,1);
	    
	    \draw[black, thin] (0.4,0,-0.5) -- +(0,2.2,0);
	    \draw[black, thin] (-0.4,0,-0.5) -- +(0,2.2,0);
	    \draw[black, thin, <->] (-0.4,2,-0.5) -- ++(0.8,0,0) node[below right, midway] {$d_m$};
	    
	    \draw[coordaxis, thin] (0,0,-0.5) -- +(1.8,0,0) node[right] {};
	    \draw[coordaxis, thin] (0,0,-0.5) -- +(0,1.8,0) node[right] {};
	    
	    \draw[red, fill=red] (0,0,-0.5) circle (0.08);
	    
	    \begin{scope}[canvas is yz plane at x=1.2]
		    \draw[coordaxis, <-] (0,-0.721) arc(-20:280:.22) node[midway, above left] {$b$};
    	\end{scope}
    	
    	\begin{scope}[canvas is xz plane at y=1.0]
		    \draw[coordaxis, <-] (0,-0.725) arc(-20:280:.2) node[midway, right] {$c$};
    	\end{scope}
	\end{scope}
\end{tikzpicture}
	\label{fig:task-space}
}
	
	\caption{Our algorithm maps a point cloud via orthographic RGBD images (a) to a $\mathit{SE}(3)$ grasp point with an additional gripper stroke $d_m$ (b). The coordinate system uses an extrinsic rotation $a$ around $z$, and then intrinsic rotations $b$ and $c$ around $x^\prime$ and $y^\prime$ respectively.}
	\label{fig:input-task-space}
\end{figure}

A grasp action $a$ is described by seven parameters: the index of the manipulation primitive $m \in \mathcal{M}$, the pose translation $(x, y, z) \in \mathbb{R}^3$ and its orientation $(a, b, c) \in [-\pi, \pi] \times [-\pi/2, \pi/2]^2$. As shown in Fig.~\ref{fig:task-space}, we define the orientation in an unusual way: The rotation $a$ around the $z$-axis is extrinsic, followed by two intrinsic rotations $b$ and $c$ around $x^\prime$ and $y^\prime$. \\

\begin{figure*}[ht]
	\centering
	\vspace{1.5mm}
	\definecolor{blueline}{rgb}{0.137,0.164,0.333}
	\definecolor{bluearea}{rgb}{0.165,0.200,0.424}

\subfloat[Collision Avoidance]{
    \begin{tikzpicture}[scale=0.72]
	    \draw[thick, blueline] (-1.5, 0) -- ++(4.5, 0) -- ++(0, 3) -- ++(1.5, 0);
	    \draw[dotted, thin, blueline, fill=bluearea!90] (-1.5, 0) -- ++(4.5, 0) -- ++(0, 3) -- ++(1.5, 0) -- ++(0, -3.15) -- ++(-6, 0) -- cycle;
	    
	    \def\angle{-31}
	
	    \draw[thick, black, fill=black!4, rotate=\angle] (-1.0, 2.5) rectangle +(2, 1.8);
	
	    \draw[red!50, dashed] (0, 0) -- +(0, 4.1);
	    \draw[red!50, dashed, rotate=\angle] (0, 0) -- +(0, 4.6);
	    \draw[<-] (0, 0) ++(60:1.5) arc (60:0:1.5);
	    
	    \node[] (angle) at (0.85, 0.32) {\small $\alpha_{min}$};
	    \draw[red, fill=red!50] (0,0) circle[radius=0.07];
	    
	    \draw[black, thin] (0, 0) -- +(0, -0.4);
	    \draw[black, thin] (3, 0) -- +(0, -0.4);
	    \draw[black, thin, <->] (0, -0.25) -- ++(3, 0) node[below, midway] {\small $r_x$};
	    
	    \draw[black, thin, white] (3, 0) -- +(0.4, 0);
	    \draw[black, thin, <->, white] (3.25, 0) -- ++(0, 3) node[right, midway] {\small $r_z$};
	    
	    \draw[black, thin, rotate=\angle] (1, 2.5) -- +(0, -0.3);
	    \draw[black, thin, rotate=\angle, <->] (0, 2.2) -- ++(1, 0) node[below left, midway] {\small $r$};
	    
	    \draw[black, thin, rotate=\angle] (0, 0) -- +(-0.3, 0);
	    \draw[black, thin, rotate=\angle, <->] (-0.2, 0) -- ++(0, 2.5) node[above left, midway] {\small $d_{l}$};
	    \draw[black, thin, rotate=\angle, <->] (-0.1, 0) -- ++(0, 4.3) node[above left, near end] {\small $d_{u}$};
    \end{tikzpicture}
}
\qquad
\subfloat[Angle $b$]{
    \begin{tikzpicture}[scale=0.72]
	    \draw[thick, blueline] (-2.5, 2.5) -- ++(1.4, 0) -- ++(0, -2.2) -- (1.3, -0.5) -- (2.5, -0.5);
	    \draw[dotted, thin, blueline, fill=bluearea!90] (-2.5, 2.5) -- ++(1.4, 0) -- ++(0, -2.2) -- (1.3, -0.5) -- (2.5, -0.5) -- ++(0, -0.15) -- ++(-5, 0) -- cycle;
	
	    \draw[thick, black, fill=black!4, rotate=-20] (-0.8, 3.6) -- ++(0, -1.6) -- ++(1.6, 0) -- ++(0, 1.6);
	    \draw[thick, black, fill=black!15, rotate=-20] (-0.2, 0.0) rectangle ++(0.4, 2.0);
	    
	    \draw[thin, green, domain=-2.5:2.5, smooth, variable=\x] plot ({\x}, {exp(-\x*\x / (2 * 0.4)) - 0.5});
	
	    \draw[red!50, dashed] (0, 0) -- +(0, 3.8);
	    \draw[red!50, dashed, rotate=-20] (0, 0) -- +(0, 3.87);
	    \draw[->] (0, 0) ++(90:2.7) arc (90:70:2.7) node[below, midway] {\small $\beta$};
	    \draw[red, fill=red!50] (0,0) circle[radius=0.07];
	    
	    \draw[white] (0, -1.0) -- (0, -1.3);
    \end{tikzpicture}
}
\qquad
\subfloat[Angle $c$]{
    \begin{tikzpicture}[scale=0.72]
	    \draw[thick, blueline, rotate=-20] (-2.5, -0.5) -- ++(1.8, 0.0) -- ++(0, 1.2) -- ++(1.5, 0) -- ++(0, -1.2) -- ++(1.5, 0);
	    \draw[dotted, thin, blueline, fill=bluearea!90, rotate=-20] (-2.5, -0.5) -- ++(1.8, 0.0) -- ++(0, 1.2) -- ++(1.5, 0) -- ++(0, -1.2) -- ++(1.5, 0) -- ++(0.05, -0.15) -- ++(-4.3, -1.5) -- cycle;
	
	    \draw[thick, black, fill=black!4, rotate=-20] (-1.4, 2.8) -- ++(0, -1.3) -- ++(2.8, 0) -- ++(0, 1.3);
	    \draw[thick, black, fill=black!15, rotate=-20] (-1.26, 0) rectangle ++(0.12, 1.5);
	    \draw[thick, black, fill=black!15, rotate=-20] (1.14, 0) rectangle ++(0.12, 1.5);
	
	    \draw[red!50, dashed] (0, 0) -- +(0, 3.2);
	    \draw[red!50, dashed, rotate=-20] (0,0) -- +(0, 3.3);
	    \draw[->] (0, 0) ++(90:2.5) arc (90:70:2.5) node[below, midway] {\small $\gamma$};
	    \draw[red, fill=red!50] (0,0) circle[radius=0.07];
	    
	    \draw[black, thin, pattern=north west lines] (1, 0.4) -- ++(0, -1.3) -- ++(-0.4, 0.15) -- cycle;
	    
	    \draw[->, rotate=-20] (-0.68, 0) -- +(-0.4, 0) node[left] {\small $\gamma_l$};
	    \draw[->] (1.0, -0.03) -- +(0.5, 0) node[right] {\small $\gamma_r$};
	    
	    \draw[->, thin, white] (0, 0) ++(90:0.5) arc (90:160:0.5);
	    \draw[->, thin, white] (0, 0) ++(90:0.5) arc (90:0:0.5);
	    
	    \draw[white] (0, -1.5) -- (0, -2);
    \end{tikzpicture}
}
	\caption{Sectional drawings of depth profiles $s(x, y)$ (blue) to illustrate the controller derivation for a given grasp point (red). In (c), the hatched area marks an undercut from the camera view from above.}
	\label{fig:angle-drawings}
\end{figure*}
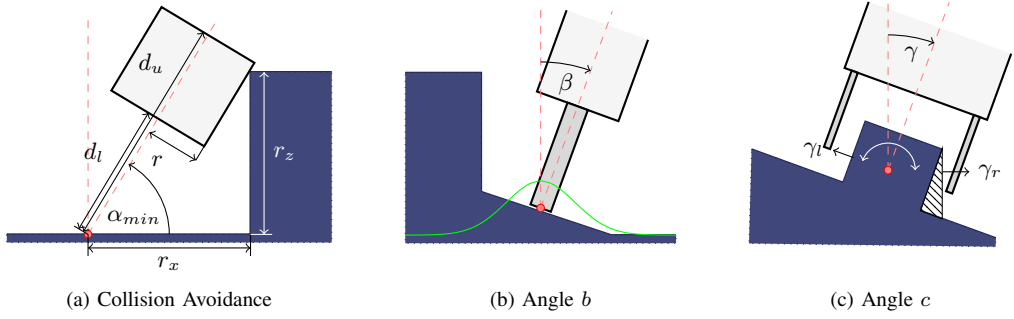

Let $s \in \mathcal{S}$ be an orthographic image of the scene. We presuppose depth information, which is also required for the orthographic projection of a recorded point cloud in real-world usage. Then, we align the task-space coordinate system with the image frame by setting it parallel to the $xy$-plane. This way, orthographic images have a key advantage over other projections: Every affine transformation of the image $s$ corresponds to a planar pose with 3\,\ac{DoF} parametrized by $(x,y,a)$. Manipulation oftentimes depends only on \textit{local} information. We simplify our approach by introducing a cropped window $s^\prime \in \mathcal{S}^\prime \subset \mathcal{S}$ at the given translation $(x, y)$ and rotation $(a)$. Furthermore, every grasp evaluation does only depend on the window $s^\prime$. Its side length corresponds roughly to the largest object size. While the intrinsic calibration of the camera needs to be known, the extrinsic calibration between camera and \ac{TCP} is learned. To fully observe a bin with high borders, we usually set the camera pose to a top-down view ($b=0, c=0$).

\subsection{Grasping Primitives}

For the set of manipulation primitives $\mathcal{M}$, we define four grasping primitives $m$ which differ in the pre-shaped gripper width $d_m$. Based on our prior work \cite{berscheid_improving_2019}, we define the general primitive as follows:
\begin{enumerate}
    \item Given a planar grasp pose $(x, y, a)$ and a gripper width $d_m$, the corresponding window area $s^\prime$ is calculated.
    \item A model-based controller takes the depth-image $s^\prime$ and outputs the height $z$ as well as the remaining 2\,\ac{DoF} for the orientation $b$ and $c$.
    \item We test for collisions and range violations and in case request a new grasp point, jumping back to (1). Using this rejection sampling approach, the model-based controller does not need to guarantee a collision-free grasp point.
    \item The robot moves according to the approach trajectory, defined as a linear motion along the $z$-axis in the local grasp frame (parallel to the gripper fingers). The gripper is pre-shaped to a width of $d_m$.
    \item If the robot detects a collision by its internal force-sensor, the robot retracts a few millimeters and closes the gripper.
    \item The object is lifted and the robot moves above the filing bin. The grasp success is then measured using the force-feedback of the gripper.
\end{enumerate}
The binary grasp reward $r$ is given by
\begin{align}
r(s, a) &= \begin{cases} 
	1 & \text{if grasp stable above filing bin,} \\
	0 & \text{else.}
\end{cases}
\end{align}
A key contribution of this work is that we see the model-based controller for $b$ and $c$ as part of the primitive, making it \textit{adaptive} based on the run-time input. The height $z$ is calculated trivially based on the center distance of the depth image. Certainly, we incorporate the model-based controller during the overall training. It therefore needs to be fixed and defined beforehand.

\subsection{Model-based Adaption}

We derive a controller $C_{bc}$
\begin{align}
    C_{bc}: \mathcal{S}^\prime \times \mathcal{M} \mapsto [-\pi / 2, \pi / 2]^2
\end{align}
mapping each image window $s^\prime$ and primitive $m$ to the angles $b$ and $c$. The \ac{FCNN} needs to implicitly estimate the controller $C_{bc}$ to predict the final grasp reward. Therefore, we constrain the mapping to be "simple" in the sense of \textit{deterministic} and \textit{continuous} with a finite and low Lipschitz-constant. As the grasp reward is generally a multi-modal distribution, a simple controller can only contribute as a first approximation. However, it suffices that the controller benefits the grasp on average, as negative (false positive) cases should be learned and eventually avoided. For an efficient implementation, we restrict our algorithms input to depth images. In comparison to planar manipulation, the robot should utilize the additional two \acp{DoF} to:
\begin{itemize}
\item Avoid \textbf{collisions} between the gripper and a possible bin or other objects.
\item Maximize the \textbf{grasp quality}. We define the quality using the antipodal score. In general, we minimize the mean scalar product of the normal vectors between both gripper jaws and the object surface.
\item Minimize \textbf{uncertainty} as grasping in 6\,\ac{DoF} naturally has to deal with back sides, undercuts and viewing shadows. In this regard, the uncertainty increases with increasing grasp angles $b$ and $c$.
\item Maximize the \textbf{grasp surface} between the gripper jaws and the object.
\end{itemize}
In a simplified manner, the grasp quality of a planar two-finger gripper is only changed by $c$. Furthermore, we separate the calculation for each \ac{DoF}.

\subsubsection{Angle $b$}
The primary task of this \ac{DoF} is to avoid collisions (Fig.~\ref{fig:angle-drawings}b). Let $s(x, y)$ be the height profile given by a depth image of size $x \in [-d_x, d_x]$ and $y \in [-d_y, d_y]$. We calculate the gradient in regard to the rotational axis $x$ and average over all values in $y$
\begin{align*}
    \tan \beta &= -\frac{1}{4 d_x d_y} \int_{-d_y}^{d_y} \int_{-d_x}^{d_x} w(x) \frac{\partial s}{\partial x}(x, y) \, \mathrm{d}x \, \mathrm{d}y
\end{align*}
The resulting angle $\beta$ approximates an orthogonal vector at the grasp point. A centered Gaussian $w(x) = \mathcal{N}(0, 1.5 \text{ finger width})$ is used as a weighting function.

\subsubsection{Angle $c$}
The primary task of this \ac{DoF} is to maximize the grasp quality score. We define the angles $\gamma_l$ and $\gamma_r$ as
\begin{align*}
    \tan \left( \gamma_{l} + \frac{\pi}{2} \right) &= -\max_y \frac{1}{2 d_x} \int_{-d_x}^{d_x} \frac{\partial s}{\partial y} (x, y)  \, \mathrm{d}x \\
    \tan \left( \gamma_{r} - \frac{\pi}{2} \right) &= -\min_y \frac{1}{2 d_x} \int_{-d_x}^{d_x} \frac{\partial s}{\partial y} (x, y)  \, \mathrm{d}x
\end{align*}
As the gradients can be quite prone to depth noise, we calculate the mean gradient
\begin{align*}
    \tan \gamma_m &= -\frac{1}{4 d_x d_y} \int_{-d_y}^{d_y} \int_{-d_x}^{d_x} \frac{\partial s}{\partial y}(x, y) \, \mathrm{d}x \, \mathrm{d}y
\end{align*}
without weights. Let $\rho \geq 0$ be the so-called \textit{certainty} parameter. We combine these three angles via
\begin{align*}
    \gamma = \begin{cases}
        \frac{1}{2}(\gamma_l + \gamma_r) + \gamma_m &\mbox{if } \gamma_l > -\frac{\pi}{2} \text{ and } \gamma_r < \frac{\pi}{2} \\
        \frac{1}{1 + \rho}(\rho \gamma_l + \gamma_m) &\mbox{if } \gamma_r \geq \frac{\pi}{2} \\
        \frac{1}{1 + \rho}(\rho \gamma_r + \gamma_m) &\mbox{if } \gamma_l \leq -\frac{\pi}{2} \\
        \gamma_m &\mbox{else}
    \end{cases}
\end{align*}
with following intuition: If an undercut exists ($\vert b \vert$ or $\vert c \vert \geq \frac{\pi}{2}$), the system will try a grasp orthogonal to the surface $\gamma_m$ (for $\rho = 0$) or antiparallel to the visible surface normal of the object $\gamma_l$ or $\gamma_r$ (for $\rho \gg 0$). Furthermore, we use $\rho = 0.5$.

\subsubsection{Collision Avoidance}
Both angles should be guaranteed to be without collision (Fig.~\ref{fig:angle-drawings}a). Therefore, we derive the interval $[\alpha_{min}, \alpha_{max}]$ around planar manipulation $\alpha=\pi/2$ without collision. For each angle, we use the maximum profile $s_b = \max_x s(x, y))$ or $s_c = \max_y s(x, y))$ regarding the rotation axis. Then, we calculate the maximum height $h_i$ at a discrete set of distances $r_{xi}$ from the grasp pose. We assume a simple rectangular collision model (with side lengths $d_{u} - d_{l}$ and $2r$) of the gripper. Simple geometric considerations lead to
\begin{align*}
    \alpha_{min} &= \begin{cases}
        \tan^{-1} \frac{r}{d_{l}} &\mbox{if } d_{u} < r_x \\
        \tan^{-1} \frac{r}{d_{u}} + \tan^{-1} \frac{\sqrt{d_{u}^2+r^2-r_x^2}}{r_x} &\mbox{if } \frac{d^2_{u} + r^2}{r_x^2 + r_z^2} < 1 \\
        \tan^{-1} \frac{r_x}{r_z} + \tan^{-1} \frac{\sqrt{r_x^2+r_z^2-r^2}}{r} + \pi &\mbox{if } \frac{d^2_{l} + r^2}{r_x^2 + r_z^2} < 1\\
        \tan^{-1} \frac{r_z}{r_x} + \tan^{-1} \frac{r}{d_{l}} &\mbox{else}
    \end{cases}
\end{align*}
depending on the finger length $d_{l}$ and the considered gripper height $d_{u}$. Similar results hold for the other side to calculate $\alpha_{max}$. Finally, the angles $b$ and $c$ are set to the clipped values of $\beta$ and $\gamma$ respectively.

\subsection{Grasp Inference}

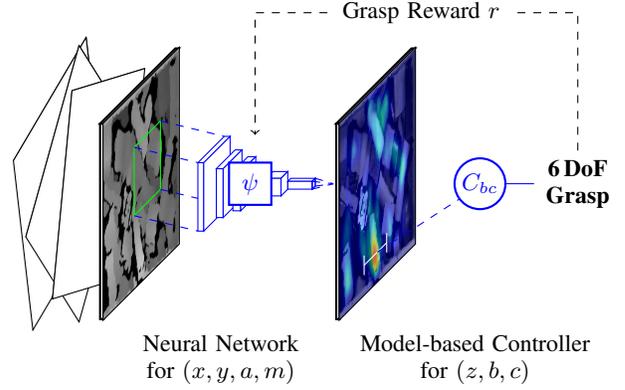
\begin{figure}[t]
    \centering
    \vspace{1mm}
\begin{tikzpicture}[font=\small, scale=0.65]
	\newcommand{\layer}[6][] {
		\draw[#1] (#2,#3 / 2,#4 / 2) -- ++(#5,0,0) -- ++(0, 0, -#4) -- ++(-#5, 0, 0) -- cycle;
		\draw[#1] (#2,#3 / 2,#4 / 2) -- ++(#5,0,0) -- ++(0, -#3, 0) -- ++(-#5, 0, 0) -- cycle;
		\draw[#1] (#2 + #5,-#3 / 2,-#4 / 2) -- ++(0,0,#4) -- ++(0,#3,0) -- ++(0,0,-#4) -- cycle;
	}
	
	\newcommand{\window}[7][] {
		\draw[#1] (#2 + #5,-#3 / 2 + #6,-#4 / 2 + #7) -- ++(0,0,#4) -- ++(0,#3,0) -- ++(0,0,-#4) -- cycle;
	}
	
	\newcommand{\calc}[7][] {
		\draw[#1, thin] (#2,-#3 / 2 + #6,-#4 / 2 + #7) -- ++(0,0,#4) -- ++(0,#3,0) -- ++(0,0,-#4) -- cycle;
		\draw[#1, thin, dashed] (#2,-#3 / 2 + #6,-#4 / 2 + #7) -- #5;
		\draw[#1, thin, dashed] (#2,#3 / 2 + #6,-#4 / 2 + #7) -- #5;
		\draw[#1, thin, dashed] (#2,-#3 / 2 + #6,#4 / 2 + #7) -- #5;
		\draw[#1, thin, dashed] (#2,#3 / 2 + #6,#4 / 2 + #7) -- #5;
	}

	\layer[fill=white, rotate around x=-75]{-1.6}{4}{4}{0}{}
	\layer[fill=white, rotate around x=-45]{-1.2}{4}{4}{0}{}
	\layer[fill=white, rotate around x=-15]{-0.8}{4}{4}{0}{}
	
	\layer[thick, fill=white]{0}{4}{4}{0.07}{}
	\begin{scope}[canvas is yz plane at x=0.07]
		\node[transform shape, rotate=-90] (a) {\includegraphics[trim=140 60 90 70, clip, width=4cm, height=4cm]{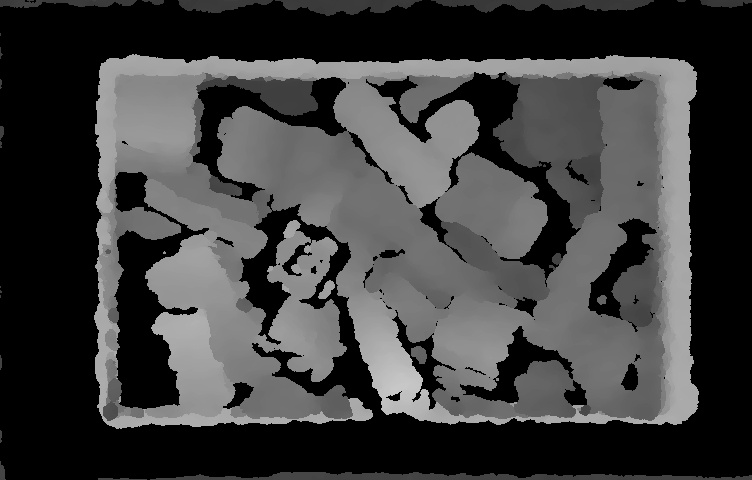}};
	\end{scope}
	
	\window[green, thin]{0}{1.4}{1.4}{0.08}{0.2}{-0.3}

	\draw[blue, thin, dashed] (0.08, -1.4 / 2 + 0.2, -1.4 / 2 - 0.3) -- (1.5, -1.4 / 2, -1.4 / 2);
	\draw[blue, thin, dashed] (0.08, -1.4 / 2 + 0.2, 1.4 / 2 - 0.3) -- (1.5, -1.4 / 2, 1.4 / 2);
	\draw[blue, thin, dashed] (0.08, 1.4 / 2 + 0.2, -1.4 / 2 - 0.3) -- (1.5, 1.4 / 2, -1.4 / 2);
	\draw[blue, thin, dashed] (0.08, 1.4 / 2 + 0.2, 1.4 / 2 - 0.3) -- (1.5, 1.4 / 2, 1.4 / 2);
		
	\layer[blue, fill=white]{1.5}{1.4}{1.4}{0.12}{}
	\layer[blue, fill=white]{1.8}{1.0}{1.0}{0.14}{}
	\layer[blue, fill=white]{2.1}{0.8}{0.8}{0.16}{}
	\layer[blue, fill=white]{2.4}{0.6}{0.6}{0.2}{}
	\layer[blue, fill=white]{2.7}{0.36}{0.36}{0.3}{}
	\layer[blue, fill=white]{3.1}{0.16}{0.16}{0.5}{}
	\calc[blue]{3.6}{0.16}{0.16}{(4.0, 0.0, 0.0)}{0}{0}
	
	\node[draw, blue, thick, minimum height=16, minimum width=16, fill=white] at (2.3, 0.0) {$\psi$};
	
	\node[text width=64, align=center] at (1.7, -3.6) {\small Neural Network \\ for $(x, y, a, m)$};
	
	\layer[thick, fill=white]{4.83}{4}{4}{0.07}{}
	\begin{scope}[canvas is yz plane at x=4.9]
		\node[transform shape, rotate=-90] (a) {\includegraphics[trim=140 60 90 70, clip, width=4cm, height=4cm]{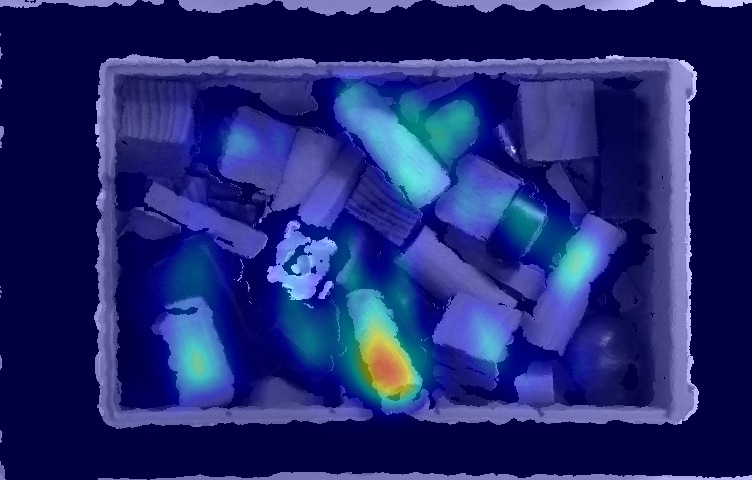}};
	\end{scope}
	
	\draw[white] (4.9, -1.4, -0.5) -- (4.9, -1.4, 0.7);
	\draw[white] (4.9, -1.2, -0.5) -- (4.9, -1.6, -0.5);
	\draw[white] (4.9, -1.2, 0.7) -- (4.9, -1.6, 0.7);
	
	\draw[blue, dashed] (4.9, -1.4, 0.0) -- (6.6, -0.3, 0.0);
	\draw[blue, thick] (7.0, 0) circle (0.52);
	\node[blue, thick] at (7.0, 0.0) {\small $C_{bc}$};
	
	\node[text width=26, align=center] (grasp) at (9.0, 0.0) {\small \textbf{6\,\ac{DoF} \\ Grasp}};
	
	\draw[blue] (7.5, 0) -- (grasp.west);
	
	\node[text width=94, align=center] at (6.9, -3.6) {\small Model-based Controller \\ for $(z, b, c)$};
	
	\node[text width=64, align=center] (grasp-reward) at (5.8, 3.5) {Grasp Reward $r$};
	\draw[dashed] (grasp.north) |- (grasp-reward.east);
	\draw[dashed, ->] (grasp-reward.west) -| (2.4, 1.0);

\end{tikzpicture}
    \caption{During inference, a \acf{FCNN} estimates the reward for a grid of grasp poses. For a single rotation and single primitive, the result can be interpreted as a reward heatmap. A model-based controller calculates the lateral \ac{DoF} $(z, b, c)$ based on the selected planar grasp $(x, y, a, m)$. During training, the reward of the \textit{adapted} grasp is fed into the \ac{FCNN}.}
    \label{fig:heatmap}
\end{figure}

Let $\psi(s^\prime)$ be a \ac{NN} mapping the image state $s^\prime$ to the single-step reward $r$. Similar to a sliding window approach, we make use of the orthographic images by evaluating $\psi$ at a discrete grid of poses. The $x$-$y$-translation is implemented efficiently as a \textit{fully-convolutional} \ac{NN}, the rotation $a$ by applying the \ac{FCNN} on multiple pre-rotated images (Fig.~\ref{fig:heatmap}). The reward for each manipulation primitive $m$ is calculated in parallel using multiple output channels of the last convolutional layer. For grasp rewards, the estimated reward $\psi$ can be interpreted as a grasp probability. \\

The overall policy $\pi(s) = \sigma \circ \psi(s)$ is composed by the reward estimation $\psi$ and a selection function $\sigma(\psi)$. During inference, a greedy selection ($\argmax$) is usually applied. Given a selected planar grasp at $(x, y, a, m)$, the height $z$ as well as the lateral angles $(b, c)$ are calculated as above.

\subsection{Self-supervised Learning}

In order to scale the real-world training, the learning process needs to work with minimal human interaction. Therefore, two bins are used for continuous and diverse training. We apply a simple  curriculum learning strategy, starting with single objects and increasing the number and complexity of object types over time. In the beginning, we sample grasps from a random selection policy and place objects in the second bin randomly. As our approach is \textit{off-policy}, we train the \ac{FCNN} using the most current dataset in parallel to the real-world data collection. Then, we sample the pick-and-place action using Boltzmann exploration, reducing the temperature $T$ over time. In the end of the training, we switch to a $\varepsilon$-greedy approach with $\varepsilon \ll 1$. During inference, we use a greedy selection or one of the best $5$ grasps if the prior one failed.

\section{EXPERIMENTAL RESULTS}

For our real-world training, a Franka Panda robot, the default gripper, and custom made jaws (Fig.~\ref{fig:overall-system}) were used. A Framos D435e RGBD-camera is mounted in the eye-in-hand configuration. Performance was measured on a system using an Intel Core i7-8700K processor and a NVIDIA GTX 1070 Ti. In front of the robot, two bins with a variety of objects for interaction are placed.

\subsection{Training Process}

\begin{figure*}[t]
	\centering
	\vspace{2mm}
	\newcommand{\subfiguresize}{0.086\textwidth}
	
\subfloat[$b$=\num{0.237}, $c$=\num{0.039}]{
	\includegraphics[width=\subfiguresize]{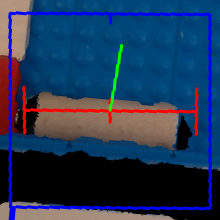}
	\includegraphics[width=\subfiguresize]{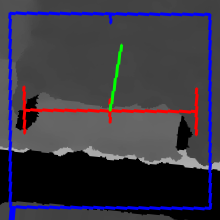}
}
~
\subfloat[$b$=\num{-0.017}, $c$=\num{-0.013}]{
	\includegraphics[width=\subfiguresize]{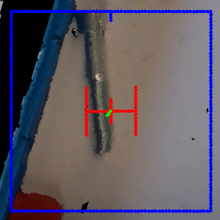}
	\includegraphics[width=\subfiguresize]{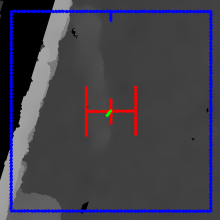}
}
~
\subfloat[$b$=\num{0.042}, $c$=\num{0.353}]{
	\includegraphics[width=\subfiguresize]{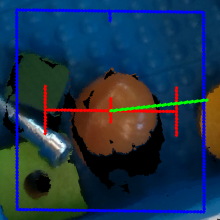}
	\includegraphics[width=\subfiguresize]{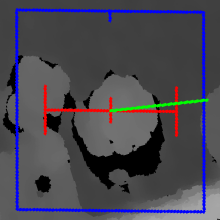}
}
~
\subfloat[$b$=\num{0.022}, $c$=\num{0.076}]{
	\includegraphics[width=\subfiguresize]{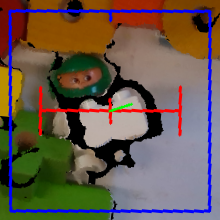}
	\includegraphics[width=\subfiguresize]{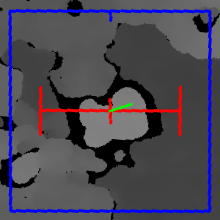}
}
~
\subfloat[$b$=\num{-0.302}, $c$=\num{0.155}]{
	\includegraphics[width=\subfiguresize]{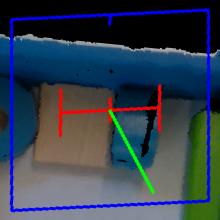}
	\includegraphics[width=\subfiguresize]{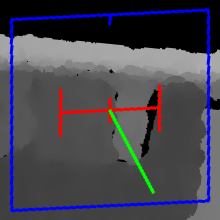}
}

\subfloat[$b$=\num{-0.002}, $c$=\num{0.281}]{
	\includegraphics[width=\subfiguresize]{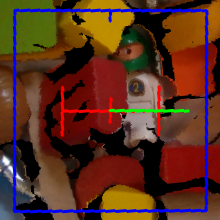}
	\includegraphics[width=\subfiguresize]{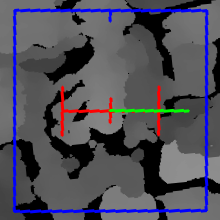}
}
~
\subfloat[$b$=\num{-0.028}, $c$=\num{-0.013}]{
	\includegraphics[width=\subfiguresize]{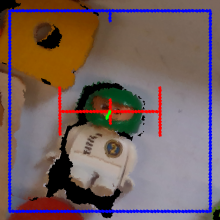}
	\includegraphics[width=\subfiguresize]{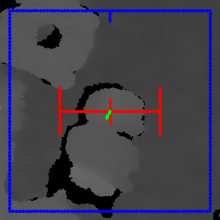}
}
~
\subfloat[$b$=\num{0.057}, $c$=\num{-0.038}]{
	\includegraphics[width=\subfiguresize]{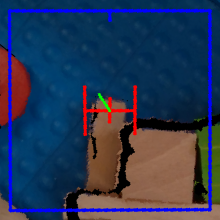}
	\includegraphics[width=\subfiguresize]{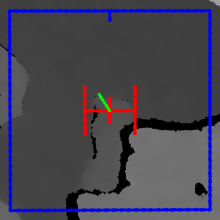}
}
~
\subfloat[$b$=\num{0.035}, $c$=\num{-0.202}]{
	\includegraphics[width=\subfiguresize]{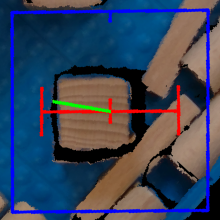}
	\includegraphics[width=\subfiguresize]{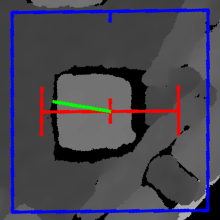}
}
~
\subfloat[$b$=\num{0.026}, $c$=\num{-0.242}]{
	\includegraphics[width=\subfiguresize]{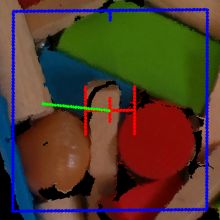}
	\includegraphics[width=\subfiguresize]{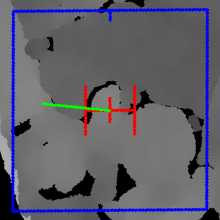}
}

	\caption{Examples of successful grasps, showing the RGBD image window $s^\prime$. The projected approach vector (green) represents the angles $b$ and $c$, the result of the model-based adaption. Furthermore, the grasp and its stroke is shown (red).}
	\label{fig:example-windows}
\end{figure*}

The primary \ac{NN} was trained with a dataset of \num{27000} grasp attempts. Typical durations for grasp attempts are around \SI{15}{s}, resulting in an overall training time of \SI{120}{h}. For exploration, random grasps were conducted for the first \num{4000} grasp attempts. Then, we apply the Boltzmann sample strategy preferring grasps with high estimated reward. The Boltzmann temperature was reduced for \num{18000} attempts, switching to a $\varepsilon$-greedy strategy with $\varepsilon = 0.1$. We retrain the \ac{NN} on the most current dataset every \num{200} grasp attempts. At the end, the \ac{BCE} of the validation dataset starts to saturate in our experimental setup. \\

The image window size is set to \num{32} $\times$ \SI{32}{px} covering a region of around \num{11} $\times$ \SI{11}{cm}. For inference, we input \num{20} rotated images of size \num{110} $\times$ \SI{110}{px}, resulting in 20 $\times$ 80 $\times$ 80 $\times$ 4 = \num{512000} actions. The \ac{FCNN} has \num{8} dilated convolutional layers with around \num{300000} parameters. Each layer follows a leaky ReLU ($\alpha=0.2$), batch normalization and dropout (between \num{0.2} and \num{0.4}). For training, the Adam optimizer with a learning rate of $10^{-4}$ was used. We augment our collected data with following techniques: (a) random color and brightness scaling, (b) random height transformation and (c) random black noise. We balance all manipulation primitives by introducing a class weight normalized by their mean reward.

\subsection{Lateral Model-based Adaption}

\begin{figure}[b!]
	\centering
\begin{tikzpicture}[scale=0.86]
\begin{axis}[
    xmin=-0.6, xmax=0.6,
	ymin=0, ymax=2100,
	ylabel=Number Grasps,
	legend pos=north east,
	width=260,
    height=130,
    xmajorgrids=true,
    ymajorgrids=true,
    xticklabels={,,}
]
	\addplot[red!70!myred, thick] table [y=n, x=angle]{figures/lateral-reward/b.txt};
	\addlegendentry{Angle $b$}

	\addplot[blue!70!myblue, thick] table [y=n, x=angle]{figures/lateral-reward/c.txt};
	\addlegendentry{Angle $c$}
\end{axis}
\end{tikzpicture}

\begin{tikzpicture}[scale=0.86]
\begin{axis}[
    xmin=-0.6, xmax=0.6,
	xlabel=Angle \lbrack rad\rbrack,
	ylabel=Grasp Success,
	width=260,
    height=120,
    xmajorgrids=true,
    ymajorgrids=true,
    y label style={yshift=0.9em},
]
    \hspace{0.118cm};
	\addplot[red!70!myred, thick] table [y=ratio, x=angle]{figures/lateral-reward/reward-b.txt};
	\addplot[blue!70!myblue, thick] table [y=ratio, x=angle]{figures/lateral-reward/reward-c.txt};
\end{axis}
\end{tikzpicture}

	\caption{Histogram of the lateral angles (top) and their corresponding mean grasp success (bottom) during training.}
	\label{fig:lateral-histogram}
\end{figure}
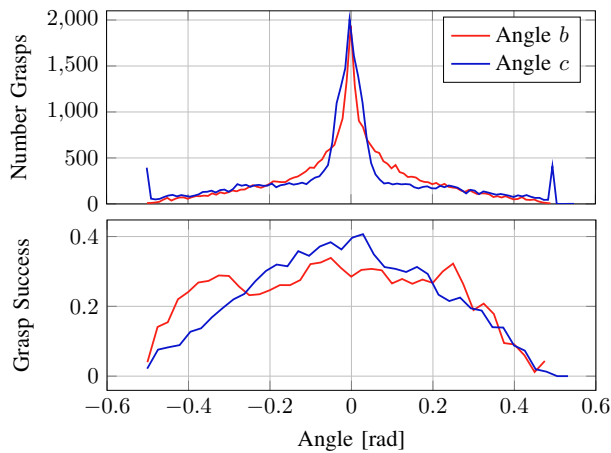

The adaption of the grasping primitive for the lateral angles $b$ and $c$ are shown for several examples in Fig.~\ref{fig:example-windows}. We find angle $b$ to be robust and working well for general collision avoidance (e.g. Fig.~\ref{fig:example-windows}a, e). By adapting $c$, the robot avoids other objects (e.g. Fig.~\ref{fig:example-windows}c, i, j) or increase the grasp quality (Fig.~\ref{fig:example-windows}f). In general, the system often uses near planar grasps (e.g. Fig.~\ref{fig:example-windows}b, g) to minimize uncertainty. Fig.~\ref{fig:lateral-histogram} shows a histogram of the angles during the training phase. As expected, we find that our model is very much limited to angles of around \SI{0.5}{rad} (\SI{30}{^\circ}). However, due to the integral limits depending on $d_m$ as well as taking $\max$ and $\min$ values, $c$ is more sensitive to camera noise. We therefore crop the lateral angles, leading to small peaks of $c$ at the border. As expected, we find that the grasp success drops off for non-planar grasps.

\subsection{Grasp Rate Evaluation}

We define the \textit{grasp rate} as the ratio of successful grasps ($r=1$) over the total number of grasp attempts without object replacement. In Table~\ref{tab:grasp-rate}, we report grasp rates for several scenarios: For trained objects, we measure grasp rates near \SI{100}{\%} for isolated objects. In bin picking scenarios (e.g. \num{20} out of \num{30} objects), the grasp rate decreases to \SI{92}{\%}. In comparison to planar grasps with a similar training, 6\,\ac{DoF} grasps are around \SI{3}{\%} less robust. If we apply our lateral adaption to a \ac{NN} trained for planar grasps, the grasp rate decreases significantly. This emphasizes the ability of the \ac{NN} to learn grasping \textit{in conjunction} with the model-based adaption. For isolated unknown objects (Fig.~\ref{fig:unknown-objects}), the robot achieves grasp rates of \SI{85}{\%}.

\begin{table}[ht]
	\centering
	\caption{Grasp rate for different bin picking scenarios: (a) 6\,\ac{DoF} or planar grasping, with (b) normal or short grippers, and (c) either trained or unknown (novel) objects. The robot needs to grasp $n$ objects out of a bin filled with $m$ objects.}
\begin{tabular}{|lr|c|c|c|}
	\hline
	& & Object Count & Grasp Rate & Number \\
	\hline
	& \multirow{4}{*}{6\,\ac{DoF}} & 1 out of 1 & \SI{100}{\%} & \num{100} \\
	& & 1 out of 5 & \SI{99.0 \pm 1.0}{\%} & \num{103} \\
	& & 5 out of 20 & \SI{96.3 \pm 1.6}{\%} & \num{109} \\
	& & 20 out of 30 & \SI{92.1 \pm 2.2}{\%} & \num{152} \\
	\hline
	\multicolumn{2}{|r|}{Planar} & 20 out of 30 & \SI{95.2 \pm 2.0}{\%} & \num{126} \\
	\multicolumn{2}{|r|}{Planar + Model} & 20 out of 30 & \SI{73.5 \pm 2.0}{\%} & \num{136} \\
	\hline
	\multirow{2}{*}{Short} & 6\,\ac{DoF} & 5 out of 20 & \SI{90.9 \pm 2.5}{\%} & \num{110} \\
	& Planar & 5 out of 20 & \SI{37.0 \pm 6.9}{\%} & \num{108} \\
	\hline
	\multirow{2}{*}{Unknown} & \multirow{2}{*}{6\,\ac{DoF}} & 1 out of 1 & \SI{85.1 \pm 3.3}{\%} & \num{215} \\
	& & 5 out of 20 & \SI{81.7 \pm 3.4}{\%} & \num{104} \\
	\hline
\end{tabular}
	\label{tab:grasp-rate}
\end{table}

To emphasize the real-world benefit of 6\,\ac{DoF} bin picking, we evaluate the grasp rate for \textit{short} gripper fingers smaller than the height of the bin. While the planar grasp rates drops naturally to below \SI{40}{\%} due to collisions, our approach keeps the grasp rate above \SI{90}{\%}. Here, the collision avoidance enables grasps that would not be possible with planar grasps. \\

For bin picking applications, inference time directly reduces the \ac{PPH}. During real-world experiments, we achieve up to \num{296 \pm 3}\,\ac{PPH}. Table~\ref{tab:inference-time} lists processing times between camera input and grasp pose output for 6\,\ac{DoF} approaches. On average, our proposed algorithm requires less than \SI{60}{ms} and scales with the observed bin area. The model-based adaption takes a constant overhead of \SI{6}{ms}.

\begin{table}[ht]
	\centering
	\caption{Processing and \ac{NN} inference time of different grasping approaches. Comparisons to related work by \cite{qin2020s4g}.}
\begin{tabular}{|r|c|c|c|}
	\hline
	& Processing [\si{ms}] & Inference [\si{ms}] & Overall [\si{ms}] \\
	\hline
	GPD\,\cite{ten2017grasp} & \num{24106} & \num{1.5} & \num{24108} \\
	Qin et al.\,\cite{qin2020s4g} & \num{5804} & \num{12.6} & \num{5817} \\
	\hline
	6\,\ac{DoF} (Ours) & \num{15.3 \pm 4.3} & \num{33.0 \pm 0.4} & \num{48.3 \pm 4.6} \\
	Planar (Ours) & \num{9.8 \pm 0.7} & \num{32.8 \pm 0.5} & \num{42.6 \pm 1.1} \\
	\hline
\end{tabular}
	\label{tab:inference-time}
\end{table}

\subsection{Auxiliary Tasks}

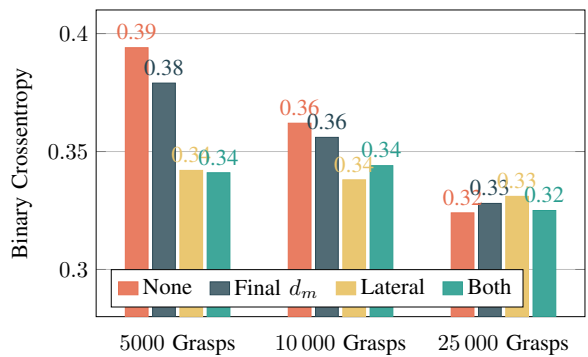
\begin{figure}[ht]
    \centering
\begin{tikzpicture}[scale=0.86]
\begin{axis}[
  xticklabels={\num{5000} Grasps, \num{10000} Grasps, \num{25000} Grasps},
  xtick=data,
  ylabel={Binary Crossentropy},
  major x tick style=transparent,
  ybar,
  ymin=0.28,
  ymax=0.41,
  width=260,
  height=180,
  xtick=data,
  ymajorgrids=true,
  nodes near coords,
  point meta=explicit symbolic,
  legend pos=south west,
  legend image code/.code={%
     \draw[#1] (0cm, -0.1cm) rectangle (0.3cm, 0.1cm);
  },
  legend style={
    legend columns=-1,
    /tikz/every even column/.append style={column sep=0.2cm}
  },
  enlarge x limits=0.25,
  x tick label style={text width=80, align=center},
]

\addplot[color=myred, fill=myred!90] coordinates {
    (1, 0.394)[0.39]
    (2, 0.362)[0.36]
    (3, 0.324)[0.32]
};

\addplot[color=myblue, fill=myblue!80] coordinates {
    (1, 0.379)[0.38]
    (2, 0.356)[0.36]
    (3, 0.328)[0.33]
};

\addplot[color=myyellow, fill=myyellow!95] coordinates {
    (1, 0.342)[0.34]
    (2, 0.338)[0.34]
    (3, 0.331)[0.33]
};

\addplot[color=mygreen, fill=mygreen!90] coordinates {
    (1, 0.341)[0.34]
    (2, 0.344)[0.34]
    (3, 0.325)[0.32]
};

\legend{None, Final $d_m$, Lateral, Both};
\end{axis}
\end{tikzpicture}
    \caption{\Acf{BCE} of the grasp reward depending on auxiliary regression tasks of the \ac{FCNN} and the training set size.}
    \label{fig:auxiliary-tasks}
\end{figure}

In general, the \ac{NN} needs to estimate the output of the model-based controller $C_{bc}$ and will implicitly learn so during the training. In this regard, it seems straight-forward to let the \ac{NN} predict the model-based lateral angles \textit{explicitly} as an auxiliary task. Moreover, we find the final (closed) gripper width $d_m$ as one more candidate that is available in real-world experiments. For both tasks, we introduce an additional regression problem, add a dedicated output to the last layer and minimize the mean square error. In Fig.~\ref{fig:auxiliary-tasks}, we evaluate the \ac{BCE} depending on auxiliary regression tasks and the training set size.
We find that in the beginning of the training, an auxiliary task can improve the \ac{BCE} up to \SI{12}{\%}. However, auxiliary losses are less useful for large dataset sizes. This corresponds to a \textit{regularizing} effect, which diminishes given enough training data.

\subsection{Camera Orientation}

\begin{figure}[t]
\centering
\vspace{2mm}
\begin{minipage}{0.5\linewidth}
  \centering
  \includegraphics[trim=0 10 340 10, clip, width=\linewidth]{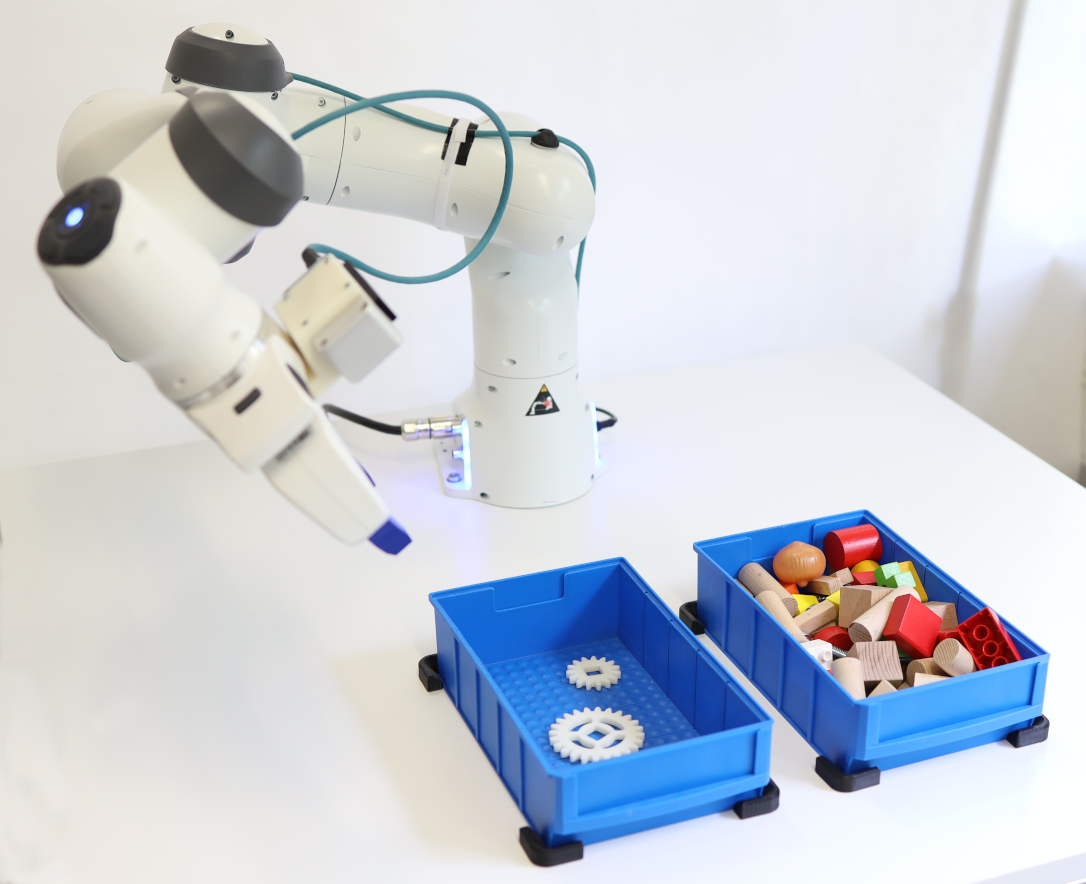}
  \captionof{figure}{Despite viewing from a lateral orientation ($\approx \SI{0.5}{rad}$), the robot is able to grasp flat objects planar to the table.}
  \label{fig:lateral-view}
\end{minipage}
~
\begin{minipage}{0.46\linewidth}
  \centering
  \includegraphics[trim=0 0 0 0, clip, width=\linewidth]{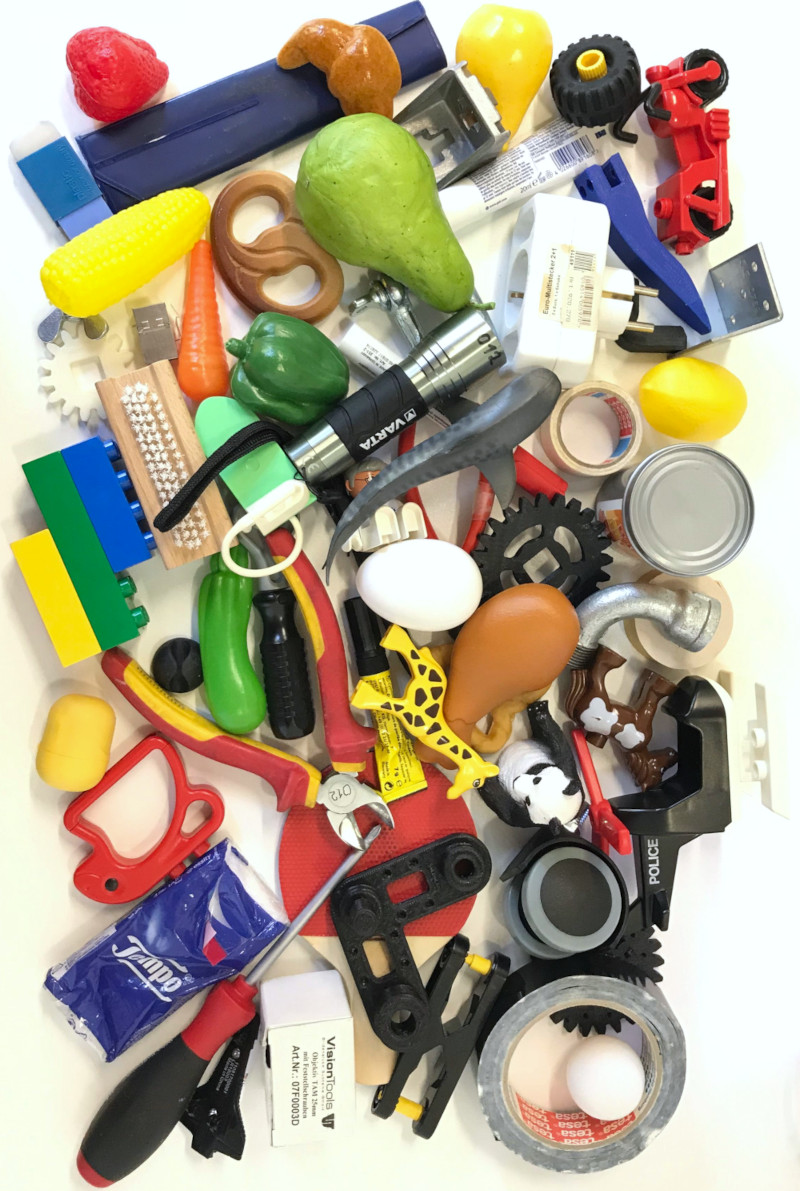}
  \captionof{figure}{Set of over \num{50} novel (unknown) objects.}
  \label{fig:unknown-objects}
\end{minipage}
\end{figure}

Training and evaluation was so far done on top-down images of the bin. For lateral camera orientations, lateral grasps improve the flexibility of the robot significantly. For example, it is now able to pick up flat objects or objects near the edge of the bin (Fig.~\ref{fig:lateral-view}).
For a camera orientation of $b=\SI{0.4}{rad}$, the system is able to pick up the shown, isolated gear in \num{26} out of \num{30} cases. In comparison, the system achieves zero successful grasps without the proposed lateral adaption.

\section{DISCUSSION AND OUTLOOK}

We have presented a hybrid approach for 6\,\ac{DoF} grasping that learns its grasp evaluation exclusively in the real world. While the planar \acp{DoF} and the gripper stroke were learned in a data-driven manner, the lateral \acp{DoF} were calculated by a model-based controller. This way, we can utilize a fully-convolutional \ac{NN} as a grasp reward estimator for an efficient implementation. Our approach achieves grasp rates of up to \SI{92}{\%} in dense clutter, despite processing times of less than \SI{50}{ms}.

In comparison to related work, we see our approach as less powerful 6\,\ac{DoF} grasping, in particular due to the lateral angle limitation of around \SI{30}{^\circ}. Our model-based controller was designed for bin picking applications in mind, in particular for collision avoidance between gripper and bin. As our grasp evaluation is learned in the real-world, we are able to avoid sim-to-real transfer in comparison to \cite{ten2017grasp, qin2020s4g, mousavian20196, schmidt2018grasping}. We find grasp rates to be sensitive to experimental details and therefore difficult to compare. In this regard, our results are similar to \cite{ten2017grasp, qin2020s4g, mousavian20196} with grasp rates above \SI{80}{\%} for unknown objects in clutter. However, we evaluate grasp rates in dense clutter and further difficult settings like short gripper fingers. As our system extends the efficient grasp evaluation based on a fully-convolutional \ac{NN}, we are able to keep the fast inference performance of planar grasping. In comparison to \cite{ten2017grasp, qin2020s4g}, our approach is around two orders of magnitude faster.

In the future, we plan to replace the model-based lateral controller with a learned model trained on analytical metrics. A hybrid actor-critic approach for the lateral angles could be used, while the planar \acp{DoF} would still be trained in the real-world. Then, the \ac{FCNN} would need to estimate the maximum reward of all lateral grasps at a given planar pose.

\bibliographystyle{IEEEtran}
\bibliography{./library}

\end{document}